\documentclass{article}

\usepackage{times}
\usepackage{graphicx} 
\usepackage{subfigure} 

\usepackage{natbib}

\usepackage{algorithm}
\usepackage{algorithmic}

\usepackage{hyperref}

\usepackage[accepted]{icml2017}
\usepackage{subfigure}

\icmltitlerunning{Feedback-Gated Rectified Linear Units}

\begin{document} 

\twocolumn[
\icmltitle{Feedback-Gated Rectified Linear Units}



\icmlsetsymbol{equal}{*}

\begin{icmlauthorlist}
\icmlauthor{Marco Kemmerling}{ma}

\end{icmlauthorlist}

\icmlaffiliation{ma}{University of Maastricht, Maastricht, The Netherlands}

\icmlcorrespondingauthor{Marco Kemmerling}{m.kemmerling@student.maastrichtuniversity.nl}

\icmlkeywords{boring formatting information, machine learning, ICML}

\vskip 0.3in
]



\printAffiliationsAndNotice{} 

\begin{abstract} 
Feedback connections play a prominent role in the human brain but have not received much attention in artificial neural network research. Here, a biologically inspired feedback mechanism which gates rectified linear units is proposed. On the MNIST dataset, autoencoders with feedback show faster convergence, better performance, and more robustness to noise compared to their counterparts without feedback. Some benefits, although less pronounced and less consistent, can be observed when networks with feedback are applied on the CIFAR-10 dataset.\end{abstract} 

\section{Introduction}
\label{introduction}
The brain has served as inspiration for artificial neural networks (ANNs) for decades. While these models are usually heavily simplified compared to the brain, they have seen significant successes in areas such as image recognition \cite{krizhevsky2012imagenet}, speech recognition \cite{hinton2012deep}, and machine translation \cite{sutskever2014sequence} in recent times. 

Despite successes, it is clear that the average human brain is vastly more powerful and versatile than any model used in practice today, and as such it may be useful to investigate how and where exactly the brain and ANNs differ. 

One such discrepancy between ANNs and the brain is the existence of feedback, or top-down connections. While there is clear evidence of prominent feedback connections in the brain, ANNs have overwhelmingly been designed based on the feedforward paradigm, although networks that do not work solely on the feedforward principle exist and are called recurrent neural networks (RNNs). Most RNNs used in practice today focus on recurrent connections from one layer to itself (e.g. LSTM networks \cite{hochreiter1997long}), which, while recurrent, arguably do not constitute top-down connections. These networks are typically applied on problems where the input consists of sequence data, where the recurrence allows for memory of previously seen elements of the sequence. 

However, the usefulness of recurrent connections or feedback is not necessarily restricted to sequence data. If the input is image data, a first look, or pass, at an image could be used to construct a rough idea of what the image contains, as well as to identify areas of interest, which can then be further examined on a second pass. 

While the network architectures considered in this paper feature real top-down connections, the focus is not on the network topology itself, but on how these top-down connections influence the behaviour of single neurons, i.e. a mechanism for incorporating feedback.

This feedback mechanism is derived from neuroscience literature and examined from two broad angles: (1) Whether the feedback mechanism can in any way improve on standard methods. Relevant metrics include convergence speed and performance quality of the trained network. (2) If examining the feedback's properties and how it behaves under certain conditions (e.g. noisy signals) can offer any insights into what role the feedback might fulfil in the brain. Needless to say, care has to be taken when trying to infer functionality of mechanisms in the brain from simplified artificial networks. Nevertheless, experimentation on artificial models offers an intriguing opportunity, as they are naturally easier to investigate and manipulate than the real brain. 

In the remainder of this paper, some neuroscientific background is explored in section \ref{neocortex} to serve as context for the feedback mechanism, followed by a description of the feedback mechanism itself as it occurs in the brain (section \ref{cellmech}). In section \ref{fgrelu} the mechanism is adapted for use in ANNs and some practical considerations on its use are given in section \ref{inpractice}. The following sections describe a range of experiments with the intention to provide answers to the research questions posed above. 

\section{Neuroscientific Background}  
\label{neocortex}
The neocortex, part of the cerebral cortex, is a part of the brain that evolved in mammals comparatively recently. It comprises around 80\% of the human brain \cite{markram2004} and is therefore often speculated to be responsible for the emergence of higher intelligence. 

The most abundant type of neuron in the neocortex is the pyramidal neuron, constituting between 70-85\% of cells. In contrast to the remaining neurons in the neocortex, so called interneurons, which are mostly inhibitory, pyramidal neurons are excitatory \cite{defelipe1992}. 

As the name suggests, pyramidal neurons have a cell body roughly shaped like a pyramid, with a base at the bottom and an apex at the top. Pyramidal neurons have two types of dendrites: basal dendrites, originating at the base, and one apical dendrite, originating at the apex. This apical dendrite terminates in what is called the apical tuft, where heavy branching of the apical dendrite occurs. \cite{defelipe1992}. 

These apical and basal dendrites are not just differently located, but also serve different functions. Basal dendrites receive regular feedforward input, while the apical tuft dendrites receive feedback input \cite{larkum2013cellular}. 

The neocortex appears to have a distinct structure which is characterised by its organisation into layers as well as columns. The columnar organisation is based on the observation that neurons stacked on top of each other tend to be connected and have similar response properties, while only few connections exist between columns. Columns are hence hypothesised to be a basic functional unit in the cortex, although this is somewhat debated in the neuroscience community \cite{goodhill2002}.

The further organisation into six layers was proposed by Brodman in 1909 \cite{brodmann1909vergleichende}. Layers 1 and 6 are of particular interest here. Layer 1 consists of almost no cell bodies, but mostly connections between axons and the apical dendrites of pyramidal neurons \cite{shipp2007}, i.e. it serves as a connection hub for feedback signals. Layer 6 sends signals to neurons in the thalamus which then in turn sends signals to layer 1 neurons in the same column \cite{shipp2007}, i.e. layers 1 and 6 create a loop where feedback is sent from layer 6 and received by layer 1.  
\subsection{Distal Input to Pyramidal Neurons}
\label{cellmech}
As described above, apical tuft dendrites receive feedback input, which appears to modulate the gain of the corresponding neuron \cite{larkum2004}. It is hypothesised that this is a way for the cortex to combine an internal representation of the world with external input, i.e. feedback to a neuron may predict whether this particular neuron should be firing, and even small feedforward input may lead the neuron to fire as long as the feedback signal is strong \cite{larkum2013cellular}. 

Taking both feedforward and feedback input into account, the firing rate of a neuron can be modelled as follows \cite{larkum2004}:

\begin{equation} 
	\label{eq:neurogain}
	f = g(\mu_S + \alpha \mu_D + \sigma + f\beta(\mu_D)-\theta)
\end{equation}

where $f$ is the firing rate of the neuron, $g$ the gain, $\mu_S$ the average somatic current (i.e. feedforward input), $\mu_D$ the average distal current (i.e. feedback input), $\alpha$ is an attenuation factor, $\sigma$ represents fluctuations in the current, $\theta$ is the firing threshold, and $\beta(\mu_D)$ is an increasing function of the dendritic mean current which saturates for values above some current threshold.  

\section{Feedback-Gated Rectified Linear Units} 
\label{fgrelu}
The model described in the previous section serves as a basis to derive an activation function which can replace the common rectified linear unit (ReLU) \cite{nair2010rectified}, i.e. $f(x) = max(0, x)$.  

To arrive at a more practical activation function, $g$ and $\theta$ are dropped from equation \ref{eq:neurogain}, since the threshold is modelled through the bias unit and the gain (i.e. slope) of a ReLU is by definition $1$ and can thus be safely dropped. Dropping the summands $\alpha \mu_D$ and $\sigma$ is less justifiable, but since they do not contribute to the core property of gain increase, they will be disregarded here, arriving at the following simplified relationship: 

\begin{equation}
	f = \mu_S + f\beta(\mu_D)
\end{equation}

Removing $f$ from the right hand side: 
\begin{equation} 
\label{eq:rate1}
	f = \frac{1}{1 - \beta(\mu_D)} \mu_S
\end{equation}

What remains is an exact definition of $\beta(\mu_D)$, which, according to \cite{larkum2004}, is ``an increasing function of the dendritic mean current $\mu$ which saturates for values above 1000pA``. In other words, the function is bounded, i.e. the gain cannot be increased to arbitrarily high values. Accordingly, some maximum value $\beta_{max}$ the function can produce and a threshold value $\eta$ which describes when this maximum is reached need to be defined. 
Assuming a piecewise linear model, $\beta(\mu_D)$ is thus defined as follows: 
\begin{equation}
\label{eq:betafunc}
	\beta(\mu_D) = min \bigg(\frac{\beta_{max}}{\eta} \mu_D, \beta_{max}\bigg)
\end{equation} 

As there are no obvious values to assign to $\beta_{max}$ and $\eta$, they are treated as hyperparameters. Since setting $\beta_{max}$ to 1 results in a division by $0$ and a value of $\beta_{max} > 1$ causes a negative slope, $\beta_{max}$ should be smaller than $1$. 

Plugging equation \ref{eq:betafunc} into equation \ref{eq:rate1} yields: 
\begin{equation}
	f = \frac{1}{1 - min(\frac{\beta_{max}}{\eta} \ \mu_D, \beta_{max})} \ \mu_S 
\end{equation}

Since negative values for $\mu_S$ are not taken into account in the above equations, $\mu_S$ is replaced with $max(0, \mu_S)$, i.e. the classic ReLU function: 
\begin{equation}
\label{eq:fgrelu}
		f = \frac{max(0, \mu_S)}{1 - min(\frac{\beta_{max}}{\eta} \ \mu_D, \beta_{max})} 
\end{equation}

\subsection{Feedback-Gated ReLUs in Practice}
\label{inpractice}
The feedback path attempts to mimic the top-down path in the brain. As such, the origin of feedback terminating in a layer should be a layer that is higher in the (feedforward) hierarchy. 

Since feedback from higher layers can only be computed if these higher layers have priorly received feedforward input, at least two time steps are needed to incorporate the modified ReLUs into a network. Concretely, some data, e.g. an image is fed into the network twice, where the first pass enables the computation of feedback which can then be utilised in the second pass. 
	Although more than two timesteps are not required, it is possible to use an arbitrary number of timesteps, which is examined in section \ref{timesteps}.
	
Any layer that receives feedback requires an additional set of weights to compute $\mu_D$. Specifically, each layer $h_i$ with size $n$ receiving feedback from layer $h_j$ with size $m$ introduces $n \times m$ additional parameters.  

The resulting networks can then be unrolled to create a feedforward network, so that for $t$ timesteps, each layer occurs $t$ times, while using the same weights at each timestep (see figure \ref{fig:autoenc}). Since the unrolled network is purely feedforward, the standard backpropagation is a suitable learning rule. 

In convolutional neural networks \cite{lecun1989generalization}, feedback is implemented on a filter-wise basis, i.e. each neuron does not receive its own unique feedback signal, but rather every filter receives a unique feedback signal that is shared between all units belonging to that filter. 	

Dropout \cite{srivastava2014dropout} should be used by dropping out the same units in all passes. Otherwise, if e.g. dropout is only applied on the last pass, the remaining units will still receive signals from dropped out units in previous passes, which defeats the purpose of dropout.

\section{Experimental Results}
\label{experiments}
The preceding sections describe a feedback mechanism and how it can be implemented in practice. Here, a range of experiments is performed to observe how this feedback mechanism changes the behaviour of ANNs. Several networks are applied on two datasets, MNIST \cite{lecun2010mnist} and CIFAR-10 \cite{krizhevsky2014cifar}. Specifically, the experiments are designed to answer the research questions posed in the introduction: (1) whether feedback can improve the performance of ANNs, (2) whether observing how the feedback works in artificial models can reveal any clues on what function feedback has in the brain. Sections \ref{constantgain}, \ref{noisyacts}, and \ref{noisycifar} serve to answer the latter question, where section \ref{constantgain} is more of a general analysis of feedback, while sections \ref{noisyacts} and \ref{noisycifar} test whether feedback might increase the networks robustness to noise. The remaining sections are concerned primarily with question (1) in that they test convergence speed and performance quality in various configurations. 

\subsection{MNIST}
\label{mnist}
The MNIST dataset is composed of $28 \times 28$ pixel binary images of handwritten digits, split into $60000$ training and $10000$ test instances \cite{lecun2010mnist}. Each image is associated with one of ten classes representing the digits between $0$ and $9$. 

The models used in the following experiments are based on a (non-convolutional) autoencoder with two encoding and two decoding layers. The input layer has dimension $(1\times784)$, the first encoding layer (E1) outputs data of dimension $(1\times392)$, the second (E2) of dimension $(1\times196)$, the first decoding layer (D1) of dimension $(1\times392)$ and the second decoding layer (D2) restores the data back to its original dimension. Except for the final layer, each layer is followed by a ReLU activation. The final layer makes use of a sigmoid activation function. 

\begin{figure}
      \centering
      \includegraphics[width=0.5\textwidth,height=5cm,keepaspectratio]{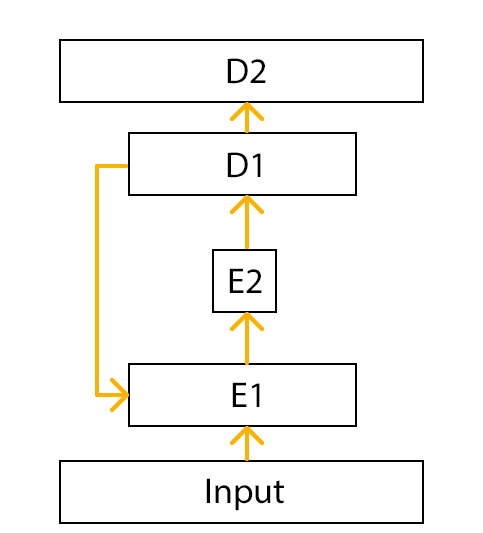}
      \includegraphics[width=0.5\textwidth,height=5cm,keepaspectratio]{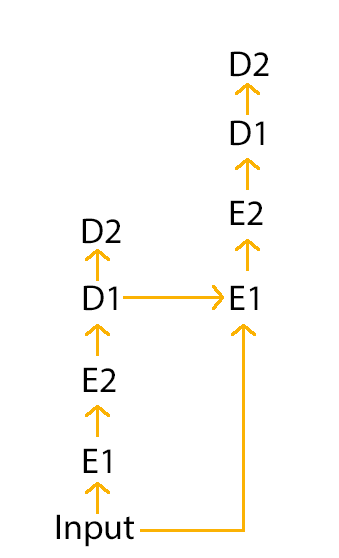}
      \caption{Left: autoencoder with (partial) feedback. Right: Unrolled autoencoder.}
      \label{fig:autoenc}
  \end{figure}

First experiments were performed with only a single feedback connection between the first decoder and the first encoder (see figure \ref{fig:autoenc}). 

Optimal values for $\eta$ and $\beta_{max}$ were determined by a grid search ($\beta_{max}=0.95, \eta=5$).
 \begin{figure}
      \centering
      \includegraphics[width=0.5\textwidth,height=5cm,keepaspectratio]{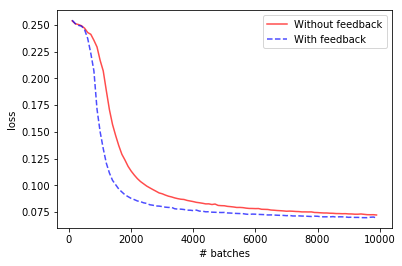}
      \caption{Test set loss of autoencoders with and without feedback. The dimension of the second encoding layer is 196. }
      \label{fig:fbvsnofb}
  \end{figure}
  
Figure \ref{fig:fbvsnofb} shows the loss curves for the autoencoder with and without feedback. While the autoencoder with feedback converges noticeably faster, the difference is relatively small. 
It is conceivable that feedback might have a greater effect if the difficulty of the task is increased. While difficulty is not a well defined term, reducing the dimension of the second encoding layer (i.e. the bottleneck) can arguably be seen as an increase in difficulty. 

\begin{figure}
      \centering
      \includegraphics[width=0.5\textwidth,height=5cm,keepaspectratio]{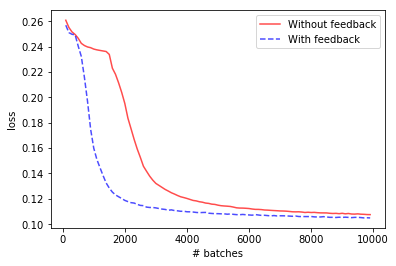}
      \caption{Test set loss of autoencoders with and without feedback. The dimension of the second encoding layer is 10. }
      \label{fig:fbvsnofb10}
  \end{figure}

The dimension of the second encoding layer is thus reduced to 10 (this modification will persist in all subsequent experiments) and the experiment is repeated. Indeed, figure \ref{fig:fbvsnofb10} shows a much larger gap between the autoencoder with feedback and the one without it, supporting the hypothesis that feedback may be more beneficial on more difficult tasks. 

\subsubsection{More Than Two Timesteps}
\label{timesteps}
While at least two timesteps are required to incorporate feedback, it is not clear whether exactly two timesteps should be used or whether $>2$ timesteps can be beneficial. To examine this, autoencoders with 1, 2, 4, 6, and 8 timesteps are trained. 

\begin{figure}
      \centering
      \includegraphics[width=0.5\textwidth,height=5cm,keepaspectratio]{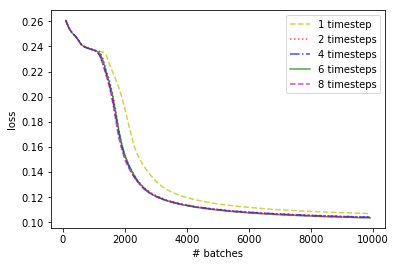}
      \caption{Autoencoder performance with varying numbers of timesteps. Each configuration was trained and evaluated 10 times. The curves shown are the averaged losses on the test set.}
      \label{fig:timesteps}
  \end{figure}

The results, depicted in figure \ref{fig:timesteps}, show that more than two timesteps yield no or negligible improvement. This may of course be data and/or task dependent. Since MNIST is a fairly simple dataset (binary images, clear separation of background and foreground, etc.), it is not inconceivable that tasks on other datasets may benefit from more than two timesteps.

\subsubsection{Comprehensive Feedback}
In the previous experiments, feedback is only sent from one decoding layer to one encoding layer. Naturally, there are many more possible configurations that incorporate further feedback connections. In the following experiment, each layer receives feedback from every layer above it, i.e. every possible top-down connection is present in the network. This will be referred to as \emph{comprehensive} feedback, whereas the previous approach will be referred to as \emph{partial} feedback. 

  \begin{figure}
      \centering
      \includegraphics[width=0.5\textwidth,height=5cm,keepaspectratio]{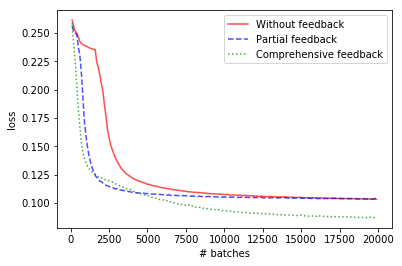}
      \caption{Loss on the test set of autoencoders without feedback, partial feedback, and comprehensive feedback. Note that the horizontal axis is different from previous figures, i.e. the training time is longer. }
      \label{fig:fullfeedback}
  \end{figure}
  
  As shown in figure \ref{fig:fullfeedback}, the configuration explained above does not only converge faster than a standard autoencoder, but also settles to a smaller loss value, which was not the case when only partial feedback was applied. 
    
 \subsubsection{Feedback vs Constant Gain} 
 \label{constantgain}
 In an effort to gain some understanding on how exactly feedback helps to improve performance, the frequency of different feedback values is examined.  A distinction is made between feedback and gain, where feedback refers to $\mu_D$ and gain refers to $\frac{1}{1 - min(\frac{\beta_{max}}{\eta} \ (\mu_D), \beta_{max})}$. 
 
\begin{figure}
      \centering
      \includegraphics[width=0.5\textwidth,height=5cm,keepaspectratio]{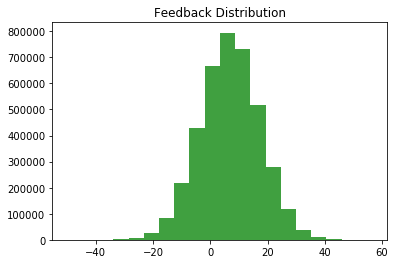}
      \includegraphics[width=0.5\textwidth,height=5cm,keepaspectratio]{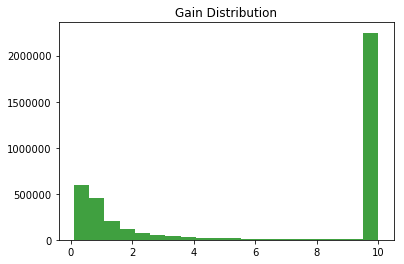}

      \caption{Distribution of feedback (top) and gain (bottom) values collected in a network with partial feedback over the complete MNIST test set. }
      \label{fig:partialhists}
  \end{figure}

 Figure \ref{fig:partialhists} shows the data as collected in a network with a single feedback connection. 
 
  While there are some smaller gain values, the overwhelming majority of values are the maximum gain the network can produce. This raises the question whether there is much benefit to learning feedback or whether it might be similarly beneficial to simply multiply all activation values by a constant. 
  
  This is easily tested by setting the gain of every ReLU in the affected layer to a constant value of 10. 
  \begin{figure}
      \centering
      \includegraphics[width=0.5\textwidth,height=5cm,keepaspectratio]{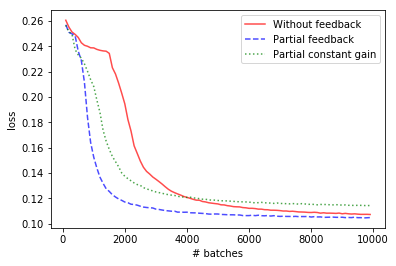}
      \caption{Comparison of a standard autoencoder, an autoencoder with partial feedback, and an autoencoder with partial constant gain (the gain of all units in the second encoding layer is set to 10)}
      \label{fig:contgainpartialloss}
  \end{figure}
  
  As can be seen in figure \ref{fig:contgainpartialloss}, this does lead to a steeper loss curve than the standard autoencoder, although not quite as steep as that of the autoencoder with actual learned feedback. Further, the performance after training is completed is worse than that of the standard autoencoder. 
  
\begin{figure}
      \centering
      \includegraphics[width=0.5\textwidth,height=5cm,keepaspectratio]{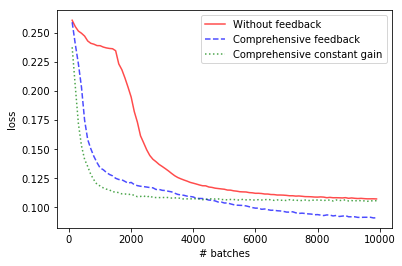}
      \caption{Comparison of a standard autoencoder, an autoencoder with comprehensive feedback, and an autoencoder with comprehensive gain (the gain of all layers is set to 10).}
      \label{fig:contgainfullloss}
  \end{figure}

  Repeating this same experiment for more than one feedback connection, i.e. for an autoencoder with comprehensive feedback, yields results as illustrated in figure \ref{fig:contgainfullloss}. 
  
  In this setup, the simple multiplication by a constant initially converges even faster than the autoencoder with learned feedback. While it does not achieve the same performance as the feedback autoencoder in later stages of training, it is on par with the standard autoencoder's performance. 
  
  Clearly, the effects of feedback cannot be fully explained by this constant gain, but the idea of a constant gain seems to have some merit.

\subsubsection{Noisy Activations} 
\label{noisyacts}
While noisy signals are usually not an issue in artificial networks, noise in the brain is very prevalent \cite{Faisal2008}. To see whether feedback makes the model more robust to noise, gaussian noise with zero mean and various standard deviations is added to the (pre-)activations of both the network with feedback and the one without it. The networks are only evaluated with added noise, training is performed without noise. Note that in the network with feedback, noise is added to the activations in both passes.

	\begin{equation}
		h = f(W^T x + b + \mathcal{N}(0,\,\sigma^{2})\,)
	\end{equation}

	As figure \ref{fig:noiseact} shows, the use of feedback significantly increases the network's robustness to noise. While this is not especially useful for machine learning models, it may be part of the reason why the feedback path exists in the brain. 

	\begin{figure}[H]
      \centering
      \includegraphics[width=0.45\textwidth,height=5cm,keepaspectratio]{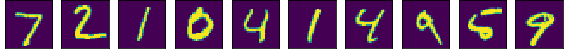}
      \includegraphics[width=0.45\textwidth,height=5cm,keepaspectratio]{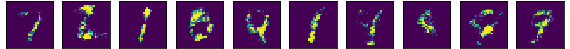}
      \includegraphics[width=0.45\textwidth,height=5cm,keepaspectratio]{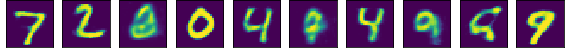}
      \caption{Gaussian noise with zero mean and standard deviation $\sigma=2.0$ is added to networks with and without feedback. The top row shows input instances to the network, the middle and bottom row show reconstructions of the network without and with feedback (respectively).}
      \label{fig:noiseact}
  \end{figure}
  
  \begin{figure}
      \centering
      \includegraphics[width=0.5\textwidth,height=5cm,keepaspectratio]{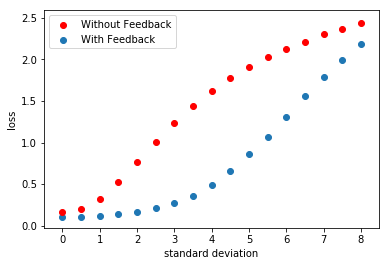}
      \caption{Gaussian noise with zero mean and varying standard deviations (horizontal) is added to networks with and without feedback. The quality of the reconstruction, as measured by the loss function (vertical axis), with respect to the magnitude of the standard deviation is shown for both networks. }
      \label{fig:noiseact}
  \end{figure}

\subsection{CIFAR-10}
The CIFAR-10 dataset is composed of $32\times32$ pixel colour images of various objects, split into 50000 training and 10000 test instances. Each image belongs to one of the following classes: airplane, automobile, bird, cat, deer, dog, frog, horse, ship, truck \cite{krizhevsky2014cifar}. 

\subsubsection{Autoencoder}
\label{cifarae}
Similarly to the MNIST experiments, an autoencoder is trained on the CIFAR-10 dataset. Again, the architecture consists of two encoding and two decoding layers. Contrary to MNIST, the encoding/decoding layers used here are convolutional/transposed convolutional layers with 16 $5 \times 5$ filters. 

\begin{figure}
      \centering
      \includegraphics[width=0.5\textwidth,height=5cm,keepaspectratio]{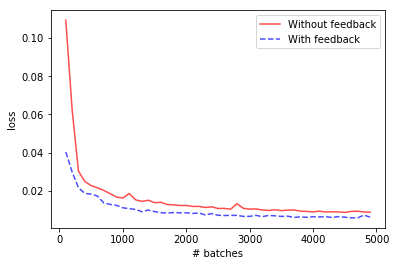}
      \caption{Test set loss of autoencoders with and without feedback on the CIFAR-10 dataset. Neither model makes use of batch normalisation. }
      \label{fig:cifarauto}
  \end{figure}

As figure \ref{fig:cifarauto} shows, the autoencoder with feedback clearly performs better than the one without it, although the difference between the two is not as pronounced as it is in the MNIST experiments.   

Curiously, if batch normalisation \cite{ioffe2015batch} is used after the activation functions, feedback cannot improve on the performance of the standard autoencoder. This may suggest that somehow feedback and batch normalisation are interacting in such a way that the feedback is rendered ineffective. 

  \begin{figure}
      \centering
      \includegraphics[width=0.5\textwidth,height=5cm,keepaspectratio]{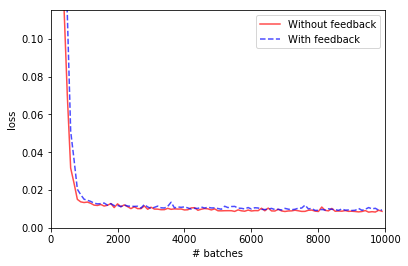}
      \caption{Test set loss of autoencoders with and without feedback on the CIFAR-10 dataset. Both models make use of batch normalisation. }
      \label{fig:cifarautobatch}
  \end{figure}

\subsubsection{Noisy Activations}
\label{noisycifar}
The experiment from section \ref{noisyacts} is repeated on the CIFAR-10 dataset. The network employed is the autoencoder without batch normalisation from the previous experiment. 

\begin{figure}
      \centering
      \includegraphics[width=0.5\textwidth,height=5cm,keepaspectratio]{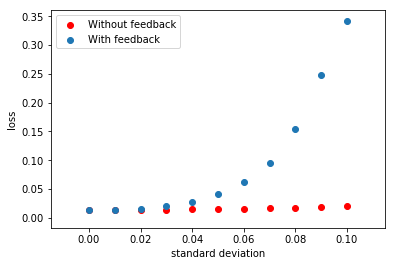}
      \caption{Gaussian noise with zero mean and varying standard deviations is added to the CIFAR-10 autoencoders with and without feedback. Although this is not apparent due to the scale of the plot, the data for the network without feedback follows a similar shape to the one with feedback.}
      \label{fig:cifarnoise}
  \end{figure}

Since feedback increased the robustness to noise in the MNIST autoencoder, the same behaviour would be expected here. However, as apparent in figure \ref{fig:cifarnoise}, the network with feedback is much more sensitive to (even small amounts of) noise than the one without feedback. 

This may be an indication that the feedback learned by the network is fundamentally different from the feedback learned in the MNIST experiments, such that it has a compounding effect on noise, rather than a rectifying one. 

\subsubsection{Classification} 
\label{cifarclass}
Classification on the CIFAR-10 dataset is performed using a convolutional neural network. The network consists of two convolutional layers with 64 filters of size $5 \times 5$, each followed by a max pooling \cite{zhou1988computation} layer with a $2\times2$ window and a stride of $2$. The convolution  and pooling layers are followed by a fully connected layer (200 units) and a softmax \cite{bridle1990probabilistic} layer. 
Batch normalisation is applied after the pooling layers and dropout with a rate of $0.5$ is applied after the pooling and the fully connected layers. 

\begin{figure}
      \centering
      \includegraphics[width=0.5\textwidth,height=5cm,keepaspectratio]{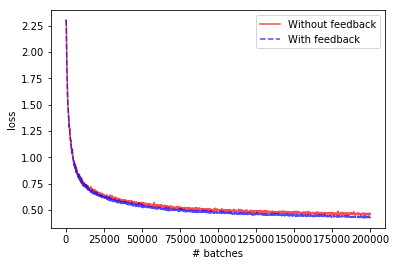}
      \caption{Classification loss on the CIFAR-10 test set. The training time of 200000 batches corresponds to 512 epochs.}
      \label{fig:cifarclass}
  \end{figure}

To test whether feedback can improve classification performance, the network is trained with (comprehensive) and without feedback. Figure \ref{fig:cifarclass} shows only a marginal performance difference between the two networks, with the feedback network being slightly better. At the end of training, the classification accuracy over the complete test set is about $0.7$\% higher for the network with feedback. 

Note that the network employed here makes use of batch normalisation, which, as shown in the previous section, may be problematic in combination with feedback. Whether this is the case here is not clear, since this particular network does not converge when batch normalisation is disabled (be it with or without feedback). 

\section{Conclusion}
The feedback mechanism presented here is able to improve performance of conventional networks both in terms of convergence speed and performance of the trained network when applied on the MNIST dataset. The benefits of feedback are less clear, however, when applied on the CIFAR-10 dataset. In principle, an autoencoder with feedback can outperform a corresponding autoencoder without feedback to a small degree, but this positive effect of feedback is negated when batch normalisation is utilised in the autoencoders. Understanding this unfavourable interaction between feedback and batch normalisation may be an opportunity to gain a deeper understanding on how feedback works and what role it fulfils. 
 
Feedback appears to have some positive effect when performing classification on CIFAR-10, although this effect is so small that drawing any firm conclusions seems ill-advised. 

When investigating the networks robustness to noise, an even larger divide between performance on MNIST and CIFAR-10 can be observed. On CIFAR-10, feedback is not only not beneficial, it actually heavily increases the network's sensitivity to noise, while the MNIST autoencoder becomes more robust when feedback is present. 

A possible explanation for this difference across datasets could be that the effectiveness of the feedback mechanism is data-dependent, i.e. it may be leveraging the highly regular structure of the MNIST dataset and is thus not as useful on the less regularly structured CIFAR-10 dataset. 

A further general difference between the experiments on the two different datasets is the use of convolutional layers, which were used in all of the CIFAR-10 experiments, but not in any of the MNIST experiments. It may be that providing feedback on a filter-wise basis is too simplistic, or that some other aspect related to convolution is not conducive to the feedback mechanism.  Further research on the combination of feedback and convolutional networks may lead to some configuration that allows for more clear benefits of feedback. 

Naturally, it might also be the case that the results on MNIST are merely an outlier, which somehow defies a more fundamental problem with the usage of feedback in current ANNs, e.g. it may be that backpropagation is not an ideal learning algorithm for feedback, or that feedback relies on more realistic models such as spiking neural networks \cite{ghosh2009spiking}.

Should clear evidence arise that feedback is useful beyond MNIST, an interesting avenue of future research would be the creation of feedback based multi-modal models, where sensory inputs from multiple different sources are combined to perform e.g. a classification task. For instance, if a network receives both visual and auditory input, the barking of a dog may result (mediated by feedback) in a higher expectation to observe a dog in the visual input. 

\section*{Acknowledgements}

I want to thank Kurt Driessens, Mario Senden, and Alexander Kroner for their supervision during this project. 

\bibliography{internship_kemmerling}

\begin{thebibliography}{21}
\providecommand{\natexlab}[1]{#1}
\providecommand{\url}[1]{\texttt{#1}}
\expandafter\ifx\csname urlstyle\endcsname\relax
  \providecommand{\doi}[1]{doi: #1}\else
  \providecommand{\doi}{doi: \begingroup \urlstyle{rm}\Url}\fi

\bibitem[Bridle(1990)]{bridle1990probabilistic}
Bridle, John~S.
\newblock Probabilistic interpretation of feedforward classification network
  outputs, with relationships to statistical pattern recognition.
\newblock In \emph{Neurocomputing}, pp.\  227--236. Springer, 1990.

\bibitem[Brodmann(1909)]{brodmann1909vergleichende}
Brodmann, Korbinian.
\newblock \emph{Vergleichende Lokalisationslehre der Grosshirnrinde in ihren
  Prinzipien dargestellt auf Grund des Zellenbaues}.
\newblock Barth, 1909.

\bibitem[DeFelipe \& Fari{\~{n}}as(1992)DeFelipe and
  Fari{\~{n}}as]{defelipe1992}
DeFelipe, Javier and Fari{\~{n}}as, Isabel.
\newblock The pyramidal neuron of the cerebral cortex: Morphological and
  chemical characteristics of the synaptic inputs.
\newblock \emph{Progress in Neurobiology}, 39\penalty0 (6):\penalty0 563--607,
  1992.

\bibitem[Faisal et~al.(2008)Faisal, Selen, and Wolpert]{Faisal2008}
Faisal, A.~Aldo, Selen, Luc P.~J., and Wolpert, Daniel~M.
\newblock Noise in the nervous system.
\newblock \emph{Nature Reviews Neuroscience}, 9\penalty0 (4):\penalty0
  292--303, 2008.

\bibitem[Ghosh-Dastidar \& Adeli(2009)Ghosh-Dastidar and
  Adeli]{ghosh2009spiking}
Ghosh-Dastidar, Samanwoy and Adeli, Hojjat.
\newblock Spiking neural networks.
\newblock \emph{International journal of neural systems}, 19\penalty0
  (04):\penalty0 295--308, 2009.

\bibitem[Goodhill \& Carreira-Perpi{\~n}{\'a}n(2002)Goodhill and
  Carreira-Perpi{\~n}{\'a}n]{goodhill2002}
Goodhill, Geoffrey~J and Carreira-Perpi{\~n}{\'a}n, Miguel~{\'A}.
\newblock Cortical columns.
\newblock \emph{Encyclopedia of cognitive science}, 2002.

\bibitem[Hinton et~al.(2012)Hinton, Deng, Yu, Dahl, Mohamed, Jaitly, Senior,
  Vanhoucke, Nguyen, Sainath, et~al.]{hinton2012deep}
Hinton, Geoffrey, Deng, Li, Yu, Dong, Dahl, George~E, Mohamed, Abdel-rahman,
  Jaitly, Navdeep, Senior, Andrew, Vanhoucke, Vincent, Nguyen, Patrick,
  Sainath, Tara~N, et~al.
\newblock Deep neural networks for acoustic modeling in speech recognition: The
  shared views of four research groups.
\newblock \emph{IEEE Signal Processing Magazine}, 29\penalty0 (6):\penalty0
  82--97, 2012.

\bibitem[Hochreiter \& Schmidhuber(1997)Hochreiter and
  Schmidhuber]{hochreiter1997long}
Hochreiter, Sepp and Schmidhuber, J{\"u}rgen.
\newblock Long short-term memory.
\newblock \emph{Neural computation}, 9\penalty0 (8):\penalty0 1735--1780, 1997.

\bibitem[Ioffe \& Szegedy(2015)Ioffe and Szegedy]{ioffe2015batch}
Ioffe, Sergey and Szegedy, Christian.
\newblock Batch normalization: Accelerating deep network training by reducing
  internal covariate shift.
\newblock In \emph{International Conference on Machine Learning}, pp.\
  448--456, 2015.

\bibitem[Krizhevsky et~al.(2012)Krizhevsky, Sutskever, and
  Hinton]{krizhevsky2012imagenet}
Krizhevsky, Alex, Sutskever, Ilya, and Hinton, Geoffrey~E.
\newblock Imagenet classification with deep convolutional neural networks.
\newblock In \emph{Advances in neural information processing systems}, pp.\
  1097--1105, 2012.

\bibitem[Krizhevsky et~al.(2014)Krizhevsky, Nair, and
  Hinton]{krizhevsky2014cifar}
Krizhevsky, Alex, Nair, Vinod, and Hinton, Geoffrey.
\newblock The cifar-10 dataset.
\newblock \emph{online: http://www. cs. toronto. edu/kriz/cifar. html}, 2014.

\bibitem[Larkum(2004)]{larkum2004}
Larkum, M.~E.
\newblock Top-down dendritic input increases the gain of layer 5 pyramidal
  neurons.
\newblock \emph{Cerebral Cortex}, 14\penalty0 (10):\penalty0 1059--1070, 2004.

\bibitem[Larkum(2013)]{larkum2013cellular}
Larkum, Matthew.
\newblock A cellular mechanism for cortical associations: an organizing
  principle for the cerebral cortex.
\newblock \emph{Trends in neurosciences}, 36\penalty0 (3):\penalty0 141--151,
  2013.

\bibitem[LeCun(1989)]{lecun1989generalization}
LeCun, Yann.
\newblock Generalization and network design strategies.
\newblock \emph{Connectionism in perspective}, pp.\  143--155, 1989.

\bibitem[LeCun et~al.(2010)LeCun, Cortes, and Burges]{lecun2010mnist}
LeCun, Yann, Cortes, Corinna, and Burges, Christopher~JC.
\newblock Mnist handwritten digit database.
\newblock \emph{AT\&T Labs [Online]. Available: http://yann. lecun.
  com/exdb/mnist}, 2, 2010.

\bibitem[Markram et~al.(2004)Markram, Toledo-Rodriguez, Wang, Gupta,
  Silberberg, and Wu]{markram2004}
Markram, Henry, Toledo-Rodriguez, Maria, Wang, Yun, Gupta, Anirudh, Silberberg,
  Gilad, and Wu, Caizhi.
\newblock Interneurons of the neocortical inhibitory system.
\newblock \emph{Nature Reviews Neuroscience}, 5\penalty0 (10):\penalty0
  793--807, 2004.

\bibitem[Nair \& Hinton(2010)Nair and Hinton]{nair2010rectified}
Nair, Vinod and Hinton, Geoffrey~E.
\newblock Rectified linear units improve restricted boltzmann machines.
\newblock In \emph{Proceedings of the 27th international conference on machine
  learning (ICML-10)}, pp.\  807--814, 2010.

\bibitem[Shipp(2007)]{shipp2007}
Shipp, Stewart.
\newblock Structure and function of the cerebral cortex.
\newblock \emph{Current Biology}, 17\penalty0 (12):\penalty0 R443--R449, 2007.

\bibitem[Srivastava et~al.(2014)Srivastava, Hinton, Krizhevsky, Sutskever, and
  Salakhutdinov]{srivastava2014dropout}
Srivastava, Nitish, Hinton, Geoffrey~E, Krizhevsky, Alex, Sutskever, Ilya, and
  Salakhutdinov, Ruslan.
\newblock Dropout: a simple way to prevent neural networks from overfitting.
\newblock \emph{Journal of machine learning research}, 15\penalty0
  (1):\penalty0 1929--1958, 2014.

\bibitem[Sutskever et~al.(2014)Sutskever, Vinyals, and
  Le]{sutskever2014sequence}
Sutskever, Ilya, Vinyals, Oriol, and Le, Quoc~V.
\newblock Sequence to sequence learning with neural networks.
\newblock In \emph{Advances in neural information processing systems}, pp.\
  3104--3112, 2014.

\bibitem[Zhou \& Chellappa(1988)Zhou and Chellappa]{zhou1988computation}
Zhou, YT and Chellappa, R.
\newblock Computation of optical flow using a neural network.
\newblock In \emph{IEEE International Conference on Neural Networks}, volume
  1998, pp.\  71--78, 1988.

\end{thebibliography}
\bibliographystyle{icml2017}

\newpage
\clearpage

\section{Appendix}

\subsection{Hyperparameter Tuning} 
As mention in section \ref{mnist}, optimal values for $\beta_{max}$ and $\eta$ are determined by a grid search. The initial grid is defined by $\eta = [5, 10, 15, \dots, 50]$ and $\beta_{max} = [0.1, 0.2, \dots, 0.8]$. 

\begin{figure}[H]
      \centering
      \includegraphics[width=0.5\textwidth,height=5cm,keepaspectratio]{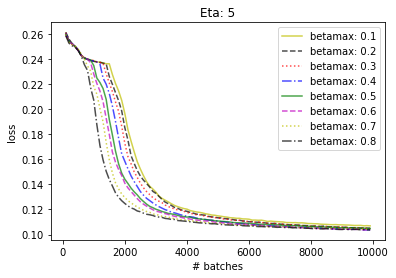}
      \includegraphics[width=0.5\textwidth,height=5cm,keepaspectratio]{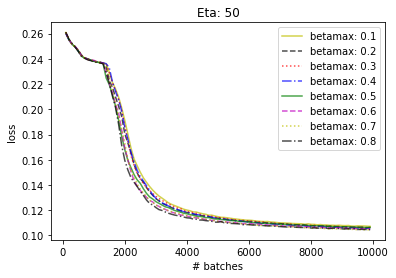}
      \caption{Autoencoder performance with varying hyperparameters. Top: $\eta$ is fixed at $5$ and $\beta_{max}$ is varied, bottom: $\eta$ is fixed at $50$ and $\beta_{max}$ is varied.}
      \label{fig:hyper1}
  \end{figure}

The highest value for $\beta_{max} \ (0.8)$ consistently shows the best performance regardless of $\eta$'s values, as exemplified by figure \ref{fig:hyper1}. Note that a high constant value of $\eta$ with varying values of $\beta_{max}$ will generally lead to less spread between the loss curves, since the activation function will be more sensitive to $\beta_{max}$ when $\eta$ is low. 

\begin{figure}[H]
      \centering
      \includegraphics[width=0.5\textwidth,height=5cm,keepaspectratio]{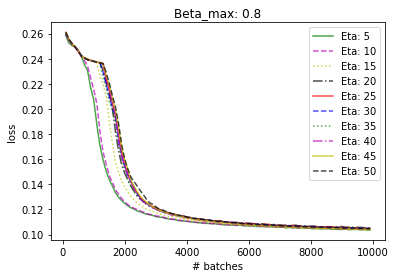}
      
      \caption{Autoencoder performance when $\beta_{max}$ is fixed at $0.8$ and $\eta$ is varied. }
      \label{fig:hyper2}
 \end{figure}

While higher values of $\beta_{max}$ lead to better performance, the inverse relationship can be seen with $\eta$, i.e. lower values of $\eta$ lead to better performance. This is illustrated in figure \ref{fig:hyper2}.  

A second grid search with $\eta = [1, 2, 3, 4, 5], \beta_{max} = [0.8, 0.85, 0.9, 0.95]$ is performed to determine whether even lower/higher values can further improve performance. Indeed, increasing $\beta_{max}$ to $0.95$ leads to better performance, but further decreasing $\eta$ is not advantegeous.

\subsection{Feedback-Controlled Threshold} 
Equation \ref{eq:neurogain} describes not only gain modulation through feedback, but also an adjustment of the activation functions threshold, i.e. $\alpha \mu_D$ is one of the terms in the summation. While gain modulation is the main property of interest in this paper, it is conceivable that the change in threshold plays a significant part in this mechanism as well. 

Incorporating this threshold mechanism into equation \ref{eq:fgrelu} leads to: 
\begin{equation}
	f = \frac{max(0, \mu_S + \alpha \mu_D)}{1 - min(\frac{\beta_{max}}{\eta} \ \mu_D, \beta_{max})} 
\end{equation}

where $\alpha$ is a parameter to be learned by the network. While $\alpha$ could also be set to a constant (tuned) value, prior experiments suggest that it is beneficial to let the network adjust $alpha$ during the course of training. 

\begin{figure}[H]
      \centering
      \includegraphics[width=0.5\textwidth,height=5cm,keepaspectratio]{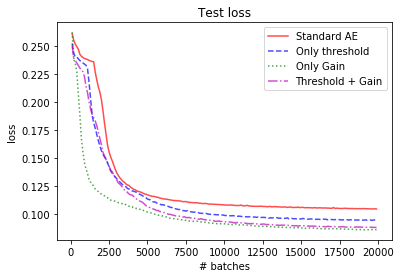}
            \caption{Performance of the standard autoencoder, an autoencoder with feedback-controlled threshold, an autoencoder with feedback-controlled gain, and an autoencoder with both feedback-controlled threshold and gain on the MNIST test set.}
      \label{fig:threshold}
  \end{figure}

As can be seen in figure \ref{fig:threshold}, the added threshold mechanism is not able to improve upon the network implementing the gain mechanism. Although the models with feedback-controlled threshold both perform better than the standard autoencoder, the model with only gain and no threshold mechanism still has the overall best performance. 

 \subsection{Input With Reduced Contrast}
Images with reduced contrast are presented to the trained (on regular contrast images) network, to see if the second pass can reconstruct an image that is more akin to a regular contrast image. To reduce the contrast, each pixel of the image is multiplied by some contrast factor $0 \leq c \leq 1$. 

Figure \ref{fig:contrastdiff} shows the absolute difference in mean pixel value between the first and second pass reconstructions for a number of different contrast factors. A high contrast input image leads to a larger difference in mean pixel value, while a low contrast image leads to a smaller difference between first and second pass reconstructions.

\begin{figure}[H]
      \centering
      \includegraphics[width=0.5\textwidth,height=5cm,keepaspectratio]{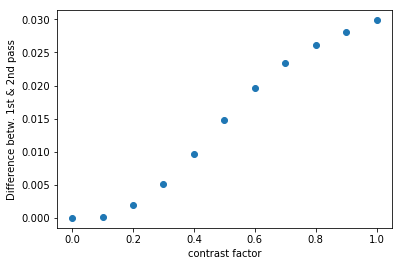}
            \caption{Absolute difference in mean pixel value between first and second pass reconstructions as a function of different contrast factors (from $0.0$ to $1.0$ in $0.1$ increments). A contrast factor of $1.0$ corresponds to no reduction in contrast, while a contrast factor of $0.0$ means the input images are entirely black.}
      \label{fig:contrastdiff}
  \end{figure}

  \begin{figure}
      \centering
      \includegraphics[width=0.5\textwidth,height=5cm,keepaspectratio]{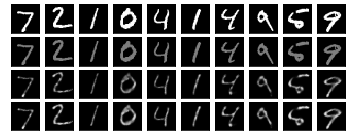}
            \caption{From top to bottom: original image, contrast reduced image, first pass reconstruction, second pass reconstruction. The contrast reduced image was produced by multiplying the original image with a contrast factor of $0.5$, i.e. each pixel in the contrast reduced image has values in the range $[0.0, 0.5]$ instead of $[0.0, 1.0]$}
      \label{fig:contrastdiff}
  \end{figure}
  
\subsection{Additional Figures} 
The following figures contain additional data that was collected as part of the experiments in section \ref{experiments}.

\vspace*{13cm}
\begin{figure}
      \centering
      \includegraphics[width=0.5\textwidth,keepaspectratio]{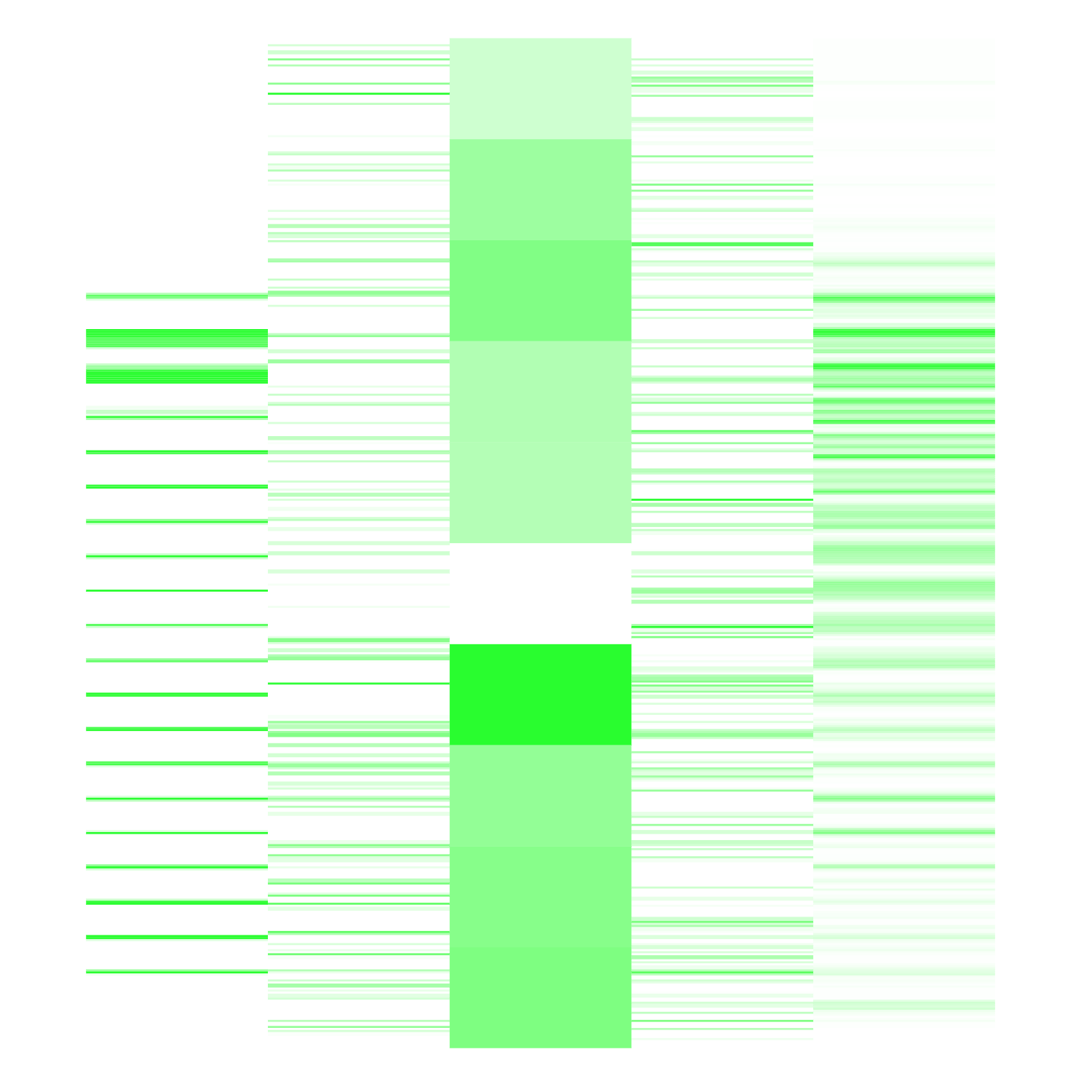}
      \includegraphics[width=0.5\textwidth,keepaspectratio]{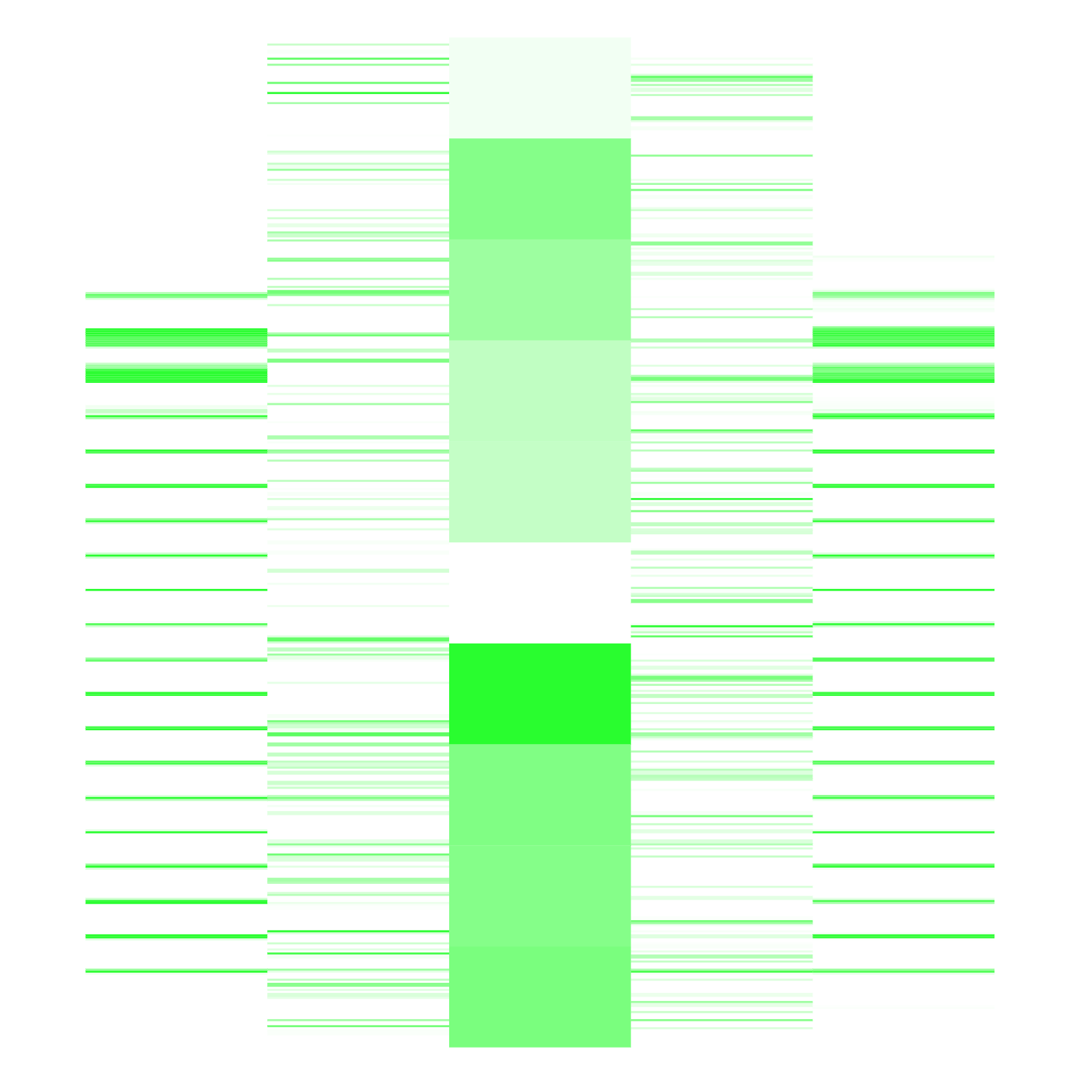}
      \caption{Visualisation of activations in the MNIST autoencoder for one particular test instance. The leftmost column corresponds to the input layer and the remaining columns correspond to the first encoding layer, the second encoding layer, the first decoding layer, and the second decoding layer, respectively. The number of rectangles in each column corresponds to the number of units in that layer. Larger values are represented by green coloured rectangles, and smaller values by white ones. Top: first pass, bottom: second pass.}
      \label{fig:actvis}
  \end{figure}

\begin{figure}
      \centering
      \includegraphics[width=0.5\textwidth,height=5cm,keepaspectratio]{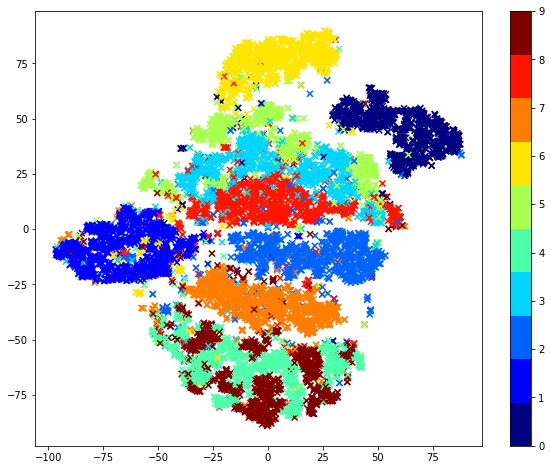}
      \includegraphics[width=0.5\textwidth,height=5cm,keepaspectratio]{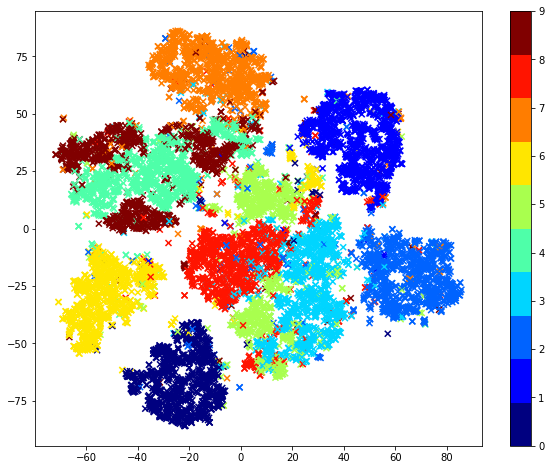}
      \includegraphics[width=0.5\textwidth,height=5cm,keepaspectratio]{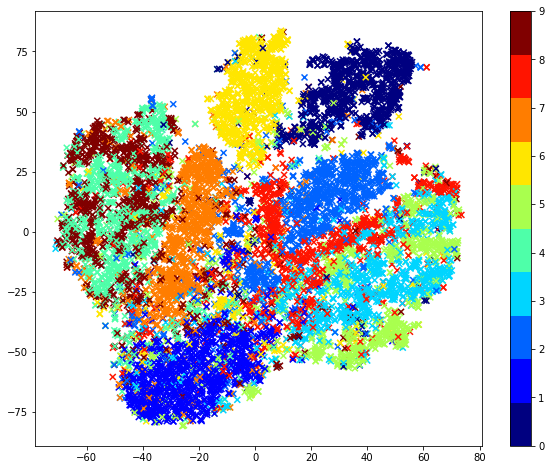}
      \includegraphics[width=0.5\textwidth,height=5cm,keepaspectratio]{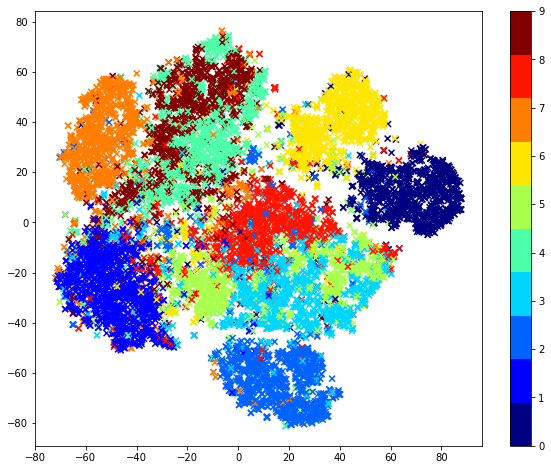}

      \caption{T-SNE visualisation of the second encoding layer of the autoencoder over the whole MNIST test set. From top to bottom: first pass, second pass, first pass with noise (as described in section \ref{noisyacts}), second pass with noise.}
      \label{fig:tsne}
  \end{figure}
%
%

  \begin{figure*}
      \centering
      \includegraphics[width=\textwidth,keepaspectratio]{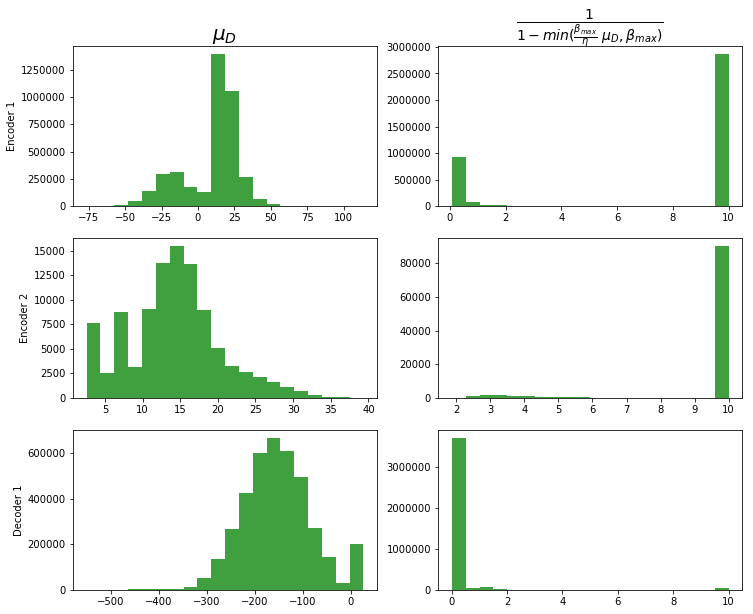}
      \caption{Histograms as seen in section \ref{constantgain}, but for the autoencoder with comprehensive feedback. }
      \label{fig:comprehensivegainhist}
  \end{figure*}
  

\begin{figure*}
  \centering
  \subfigure{\includegraphics[width=0.3\textwidth,keepaspectratio]{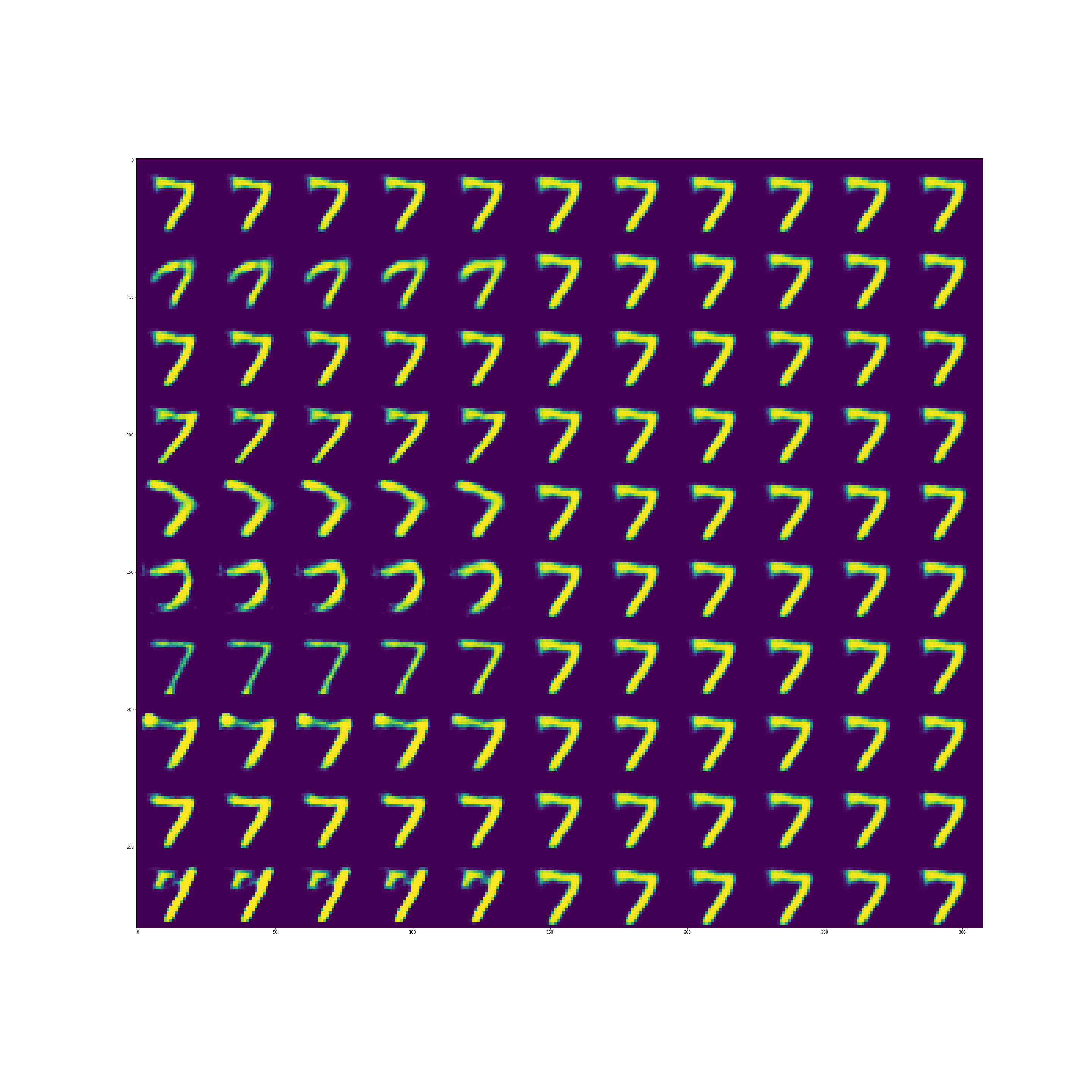}
      \includegraphics[width=0.3\textwidth,keepaspectratio]{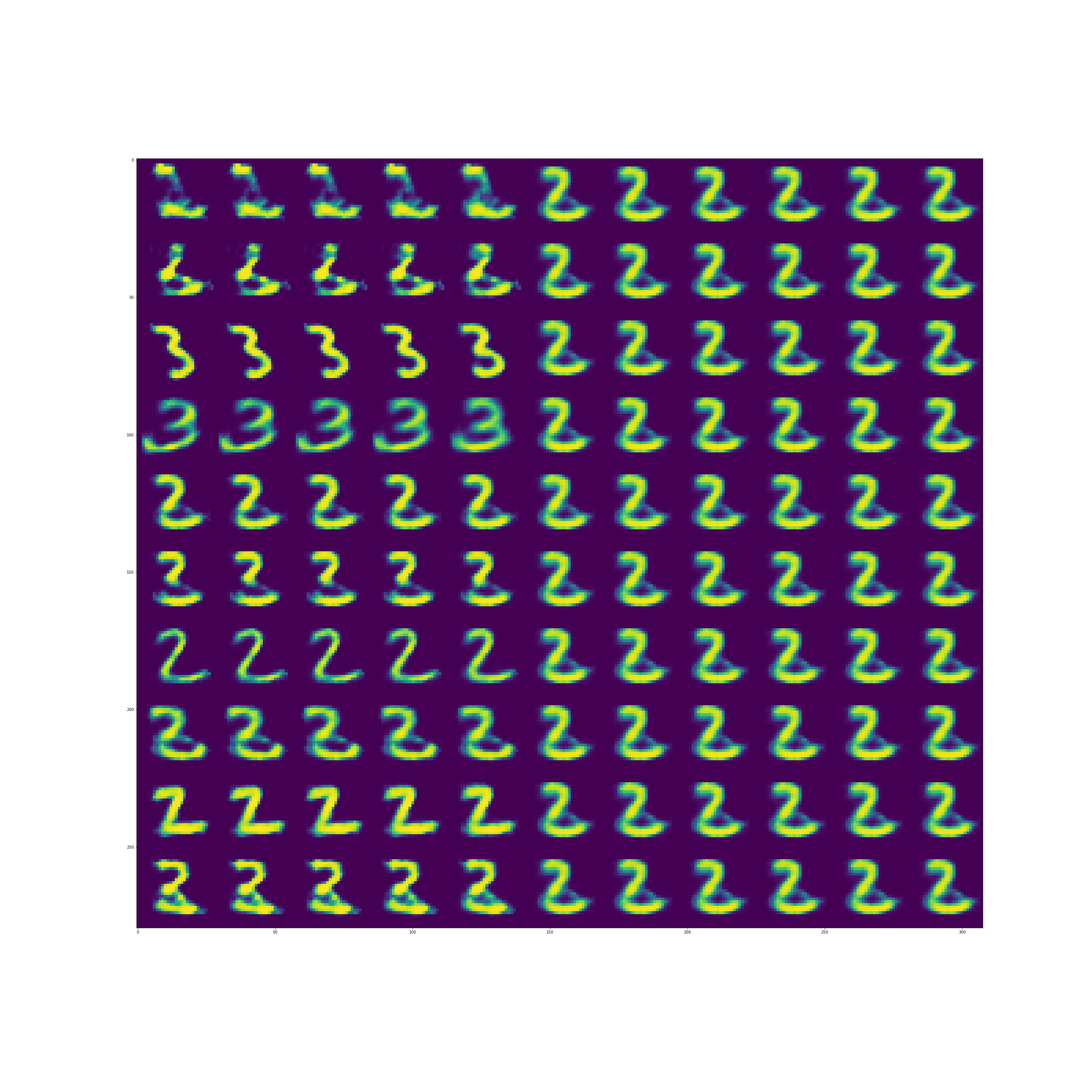}
      \includegraphics[width=0.3\textwidth,keepaspectratio]{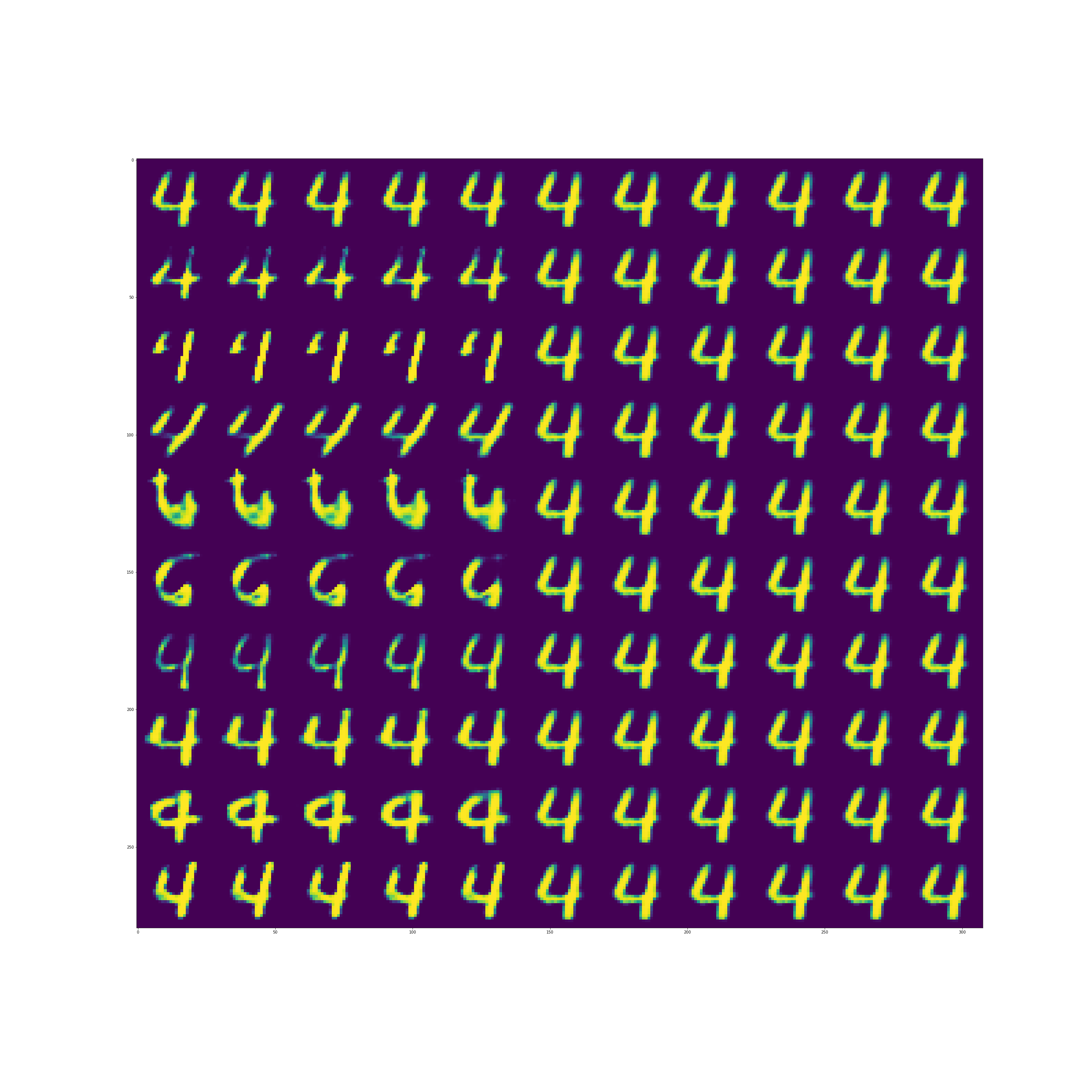}
      
}
\subfigure{\includegraphics[width=0.3\textwidth,keepaspectratio]{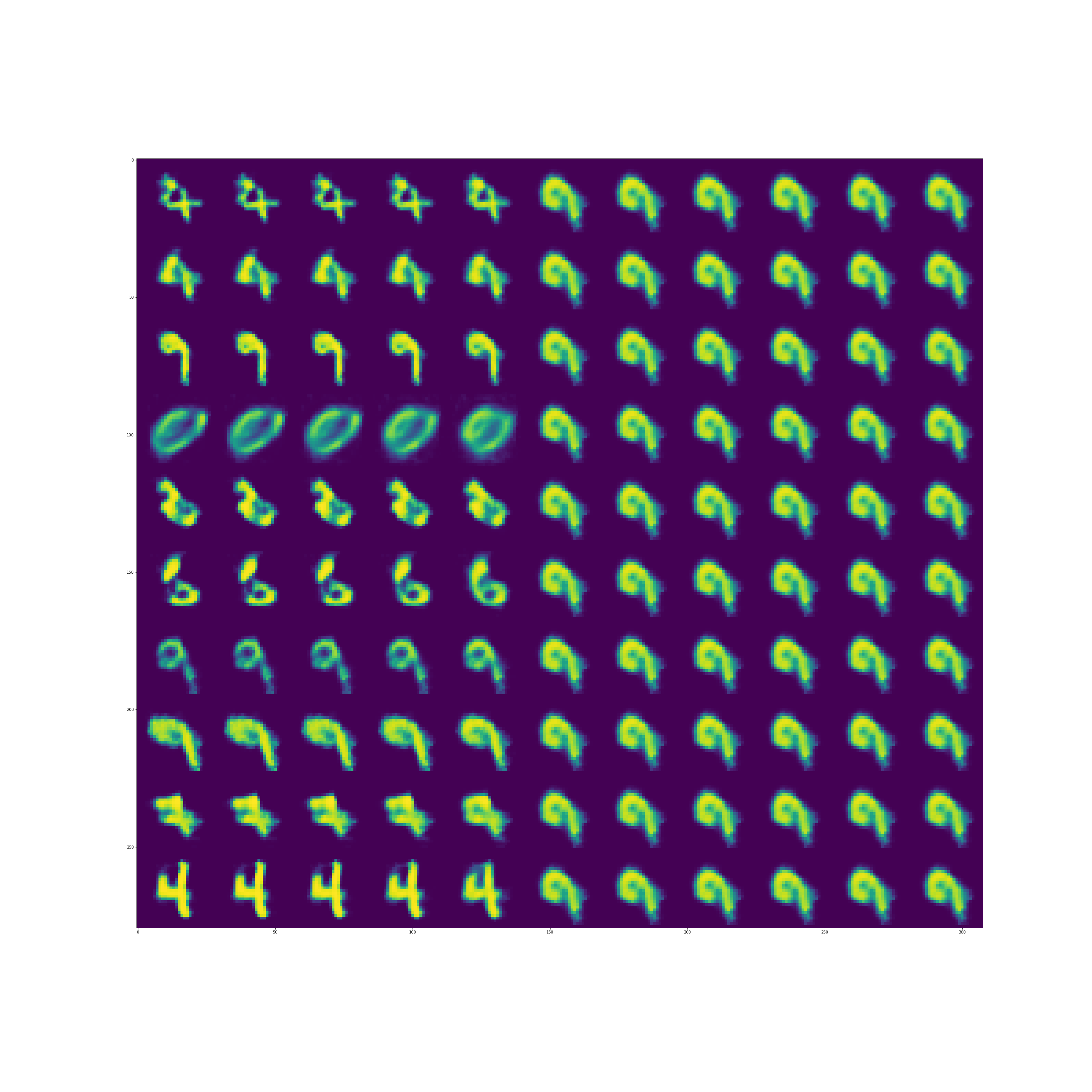}

      \includegraphics[width=0.3\textwidth,keepaspectratio]{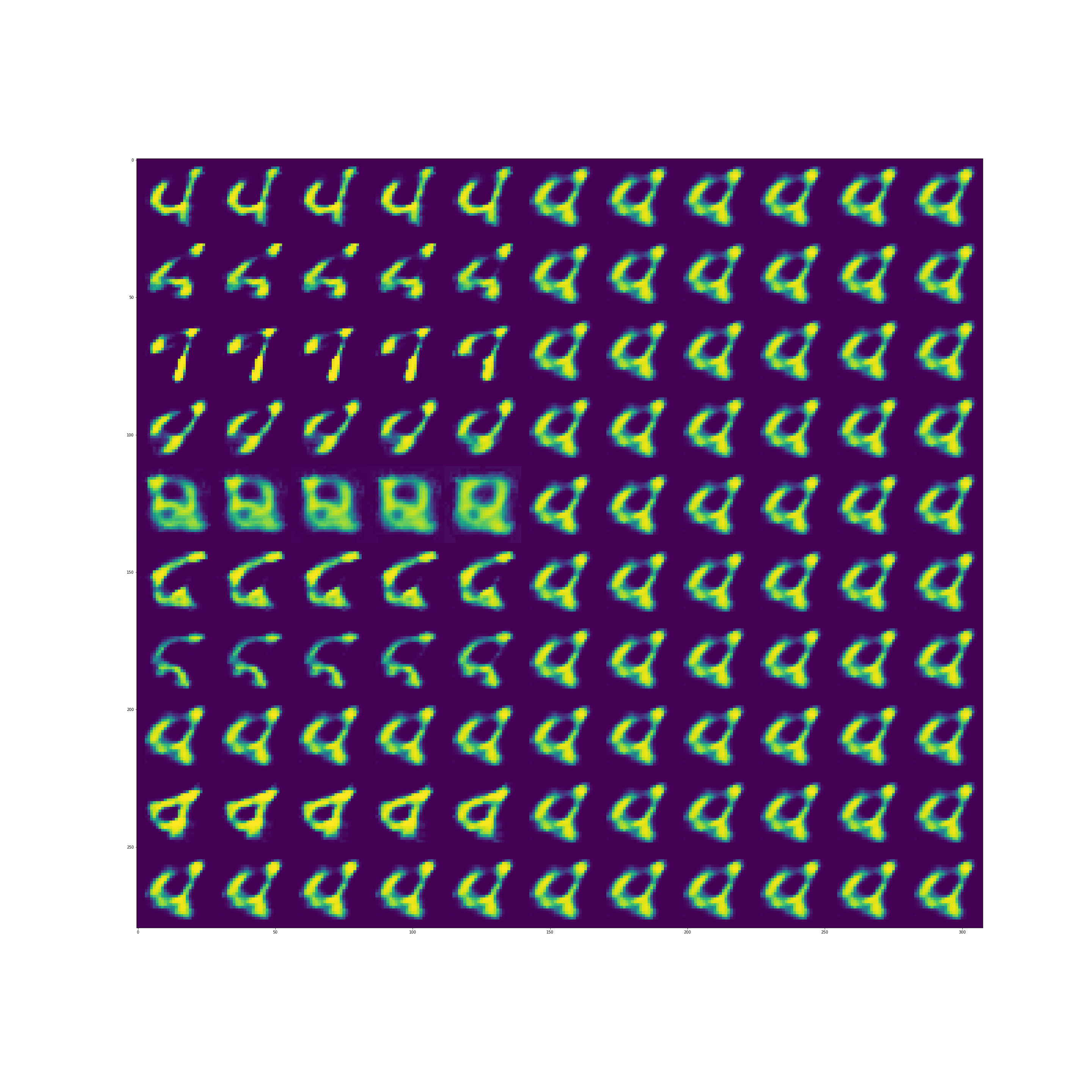}
      \includegraphics[width=0.3\textwidth,keepaspectratio]{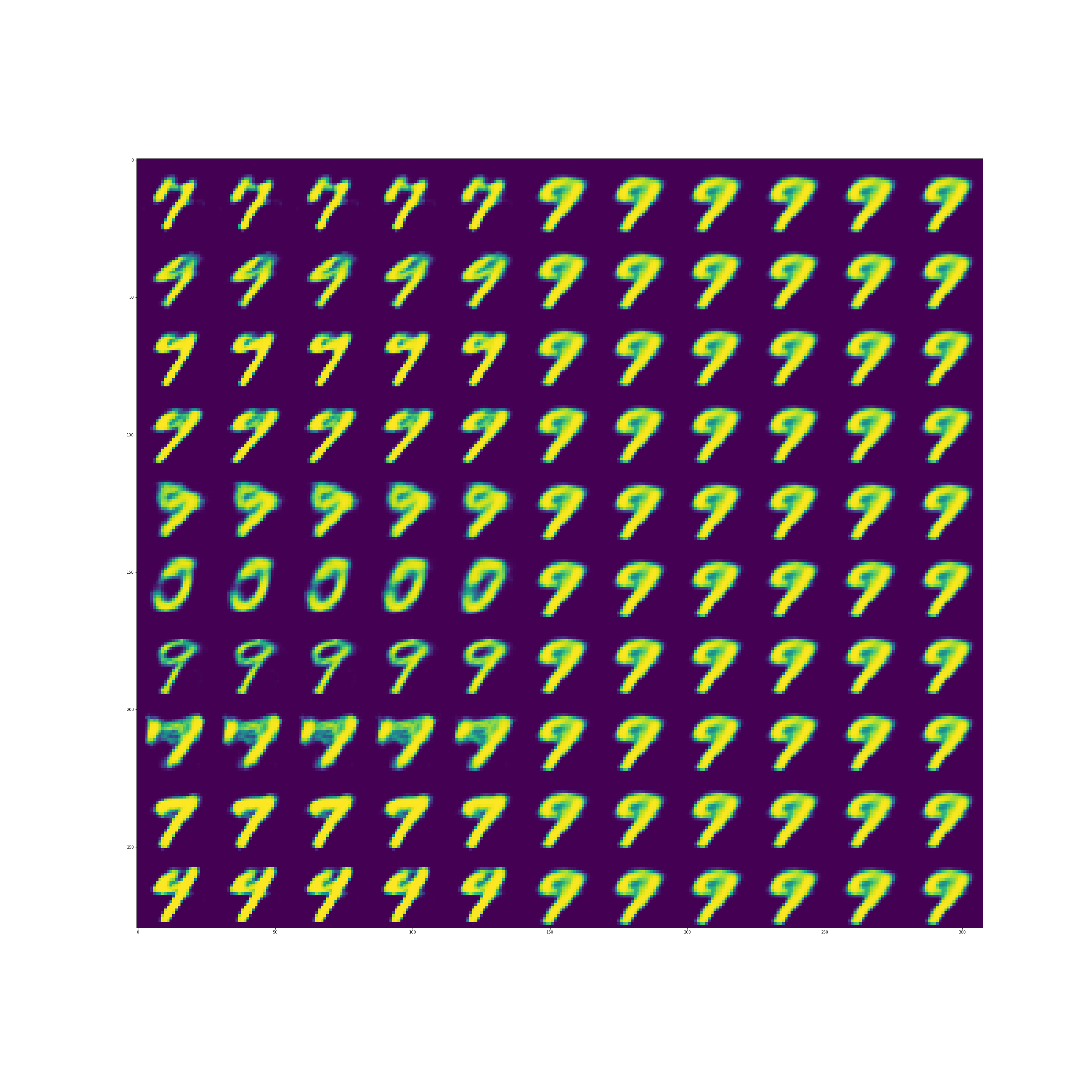}

}\quad
  \subfigure{ \includegraphics[width=0.3\textwidth,keepaspectratio]{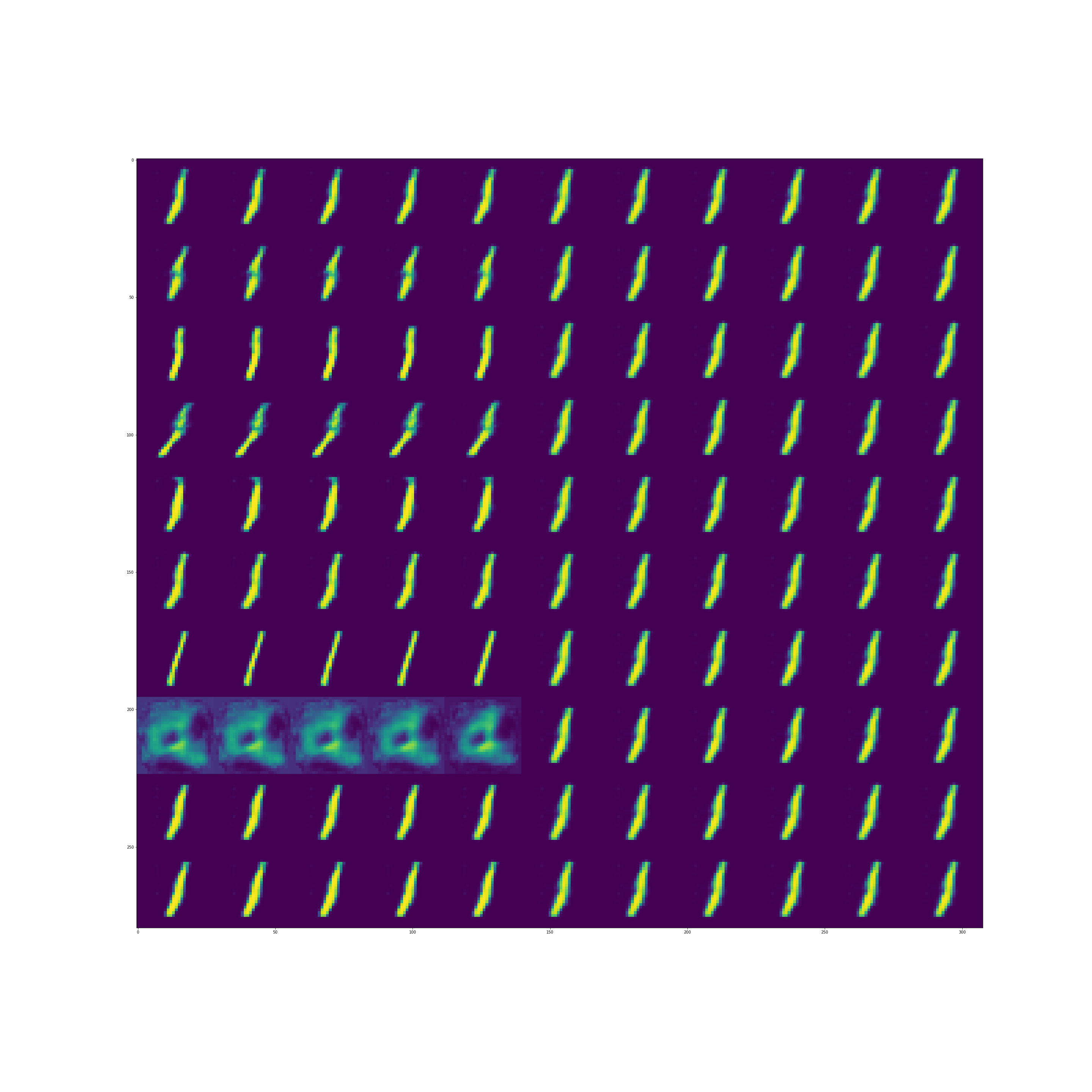}

     \includegraphics[width=0.3\textwidth,keepaspectratio]{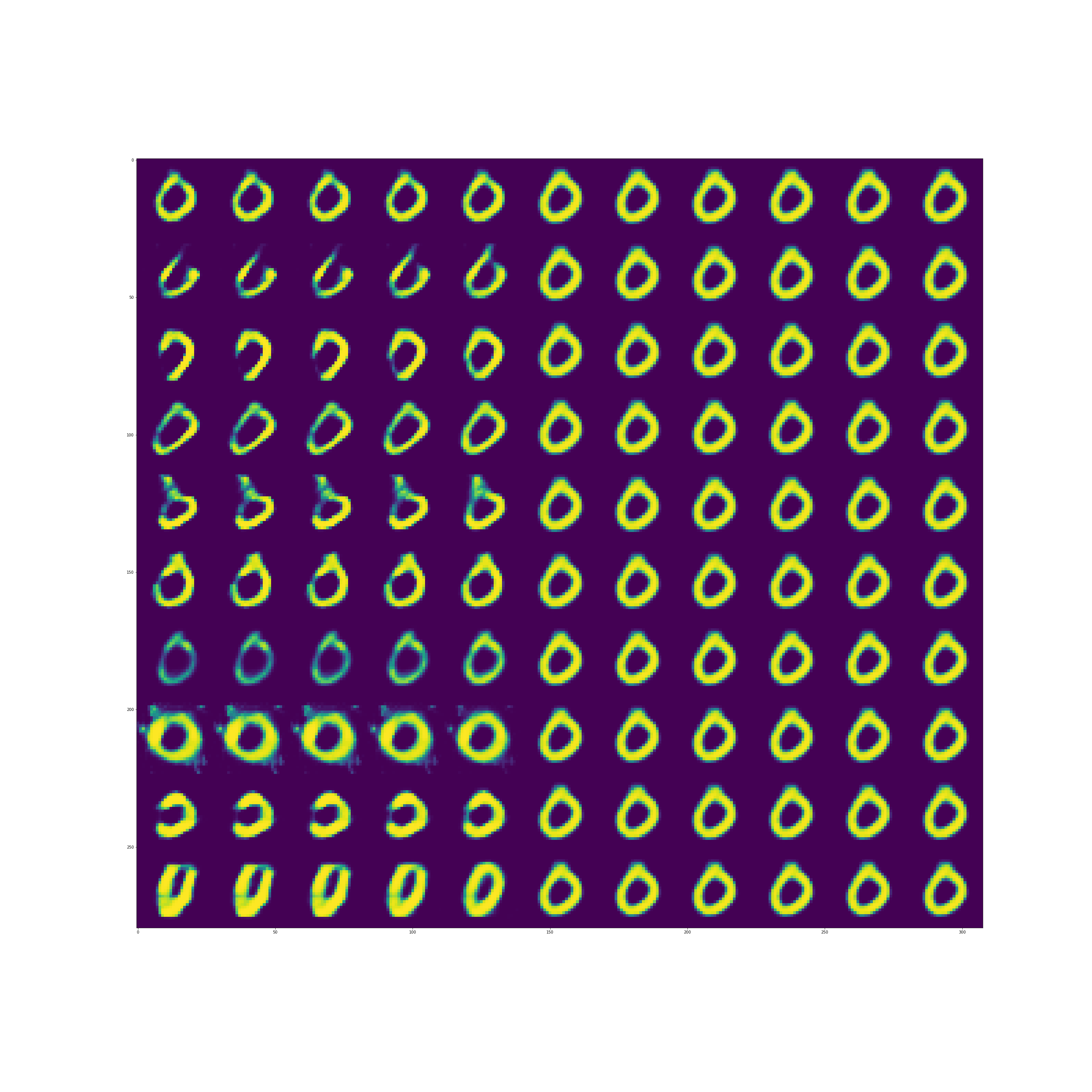}
     \includegraphics[width=0.3\textwidth,keepaspectratio]{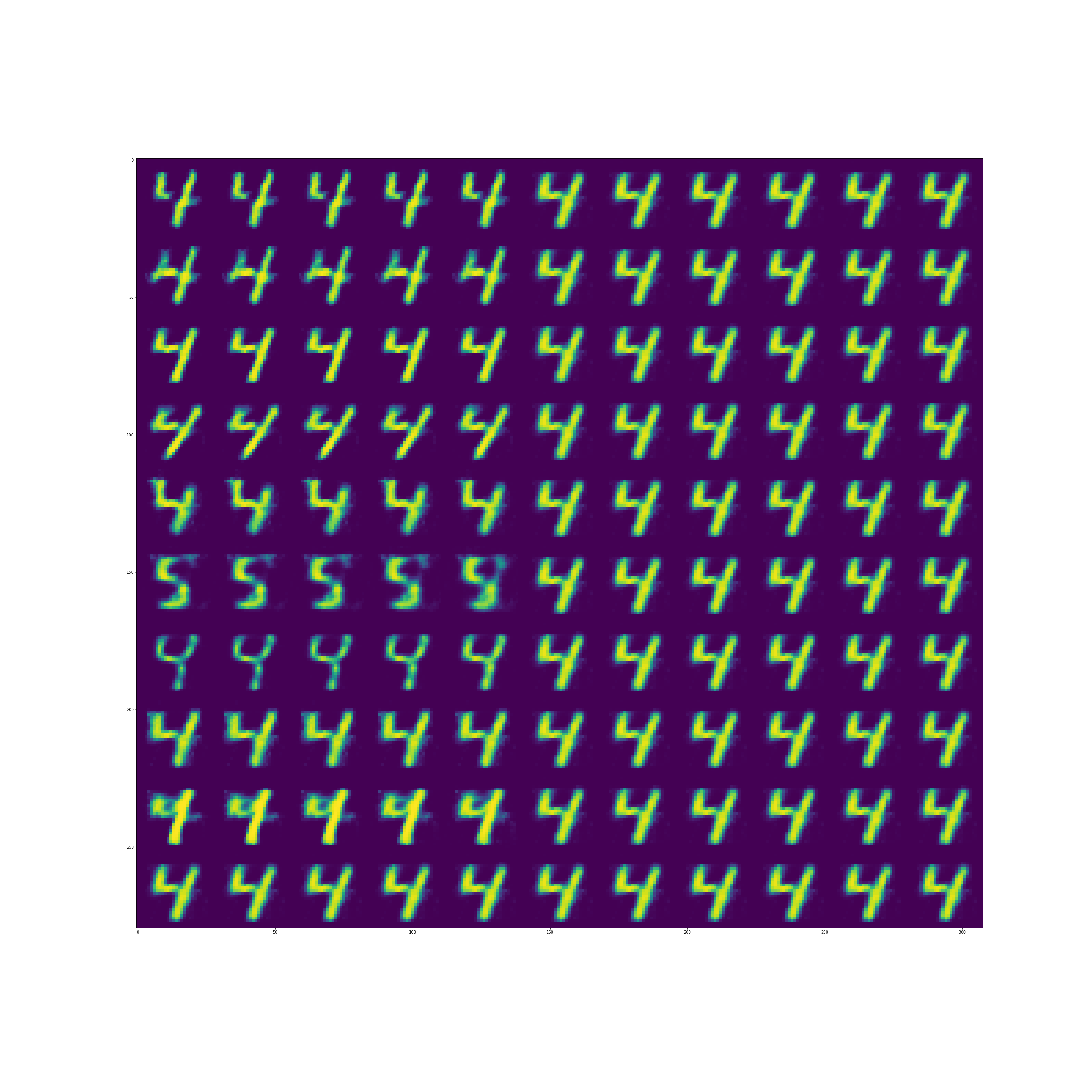}
	}
	
\caption{Different gain values are manually fed into the second encoding layer of the network and the resulting reconstruction is visualised. In each of the above images, one specific input image is presented to the network, but the gain is varied. In row $i$ of each image, every unit of the second encoding layer receives a gain of 10, except for unit $i$, which receives a gain between $0$ and $10$, depending on the column it is in. When using the MNIST autoencoder with comprehensive feedback, it can be observed that only one unit in the second encoding layer has any variation in gain (the remaining ones have a constant gain of 10 regardless of the input). This one unit corresponds to the fourth row from the bottom of each image and seems to be responsible for setting the `intensity` of the reconstruction.}
\end{figure*}

\begin{figure}
      \centering
      \includegraphics[width=\textwidth,height=5cm,keepaspectratio]{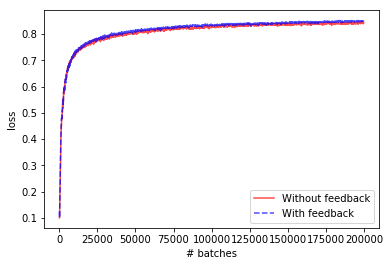}
      
      \includegraphics[width=\textwidth,height=5cm,keepaspectratio]{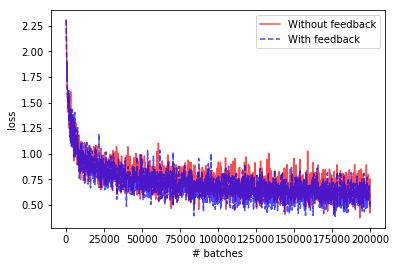}

	\includegraphics[width=\textwidth,height=5cm,keepaspectratio]{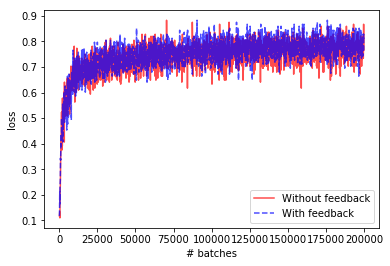}
	
	\includegraphics[width=\textwidth,height=5cm,keepaspectratio]{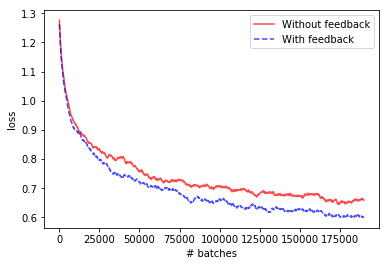}

            \caption{CIFAR-10 classification as seen in section \ref{cifarclass}. From top to bottom: test set accuracy, training set loss, training set accuracy, training set loss after applying a moving average filter (window size 100). }
      \label{fig:addcifarclass}
  \end{figure}


\end{document}